\documentclass[sn-mathphys]{sn-jnl}

\jyear{2021}%

\theoremstyle{thmstyleone}%
\theoremstyle{thmstyletwo}%

\theoremstyle{thmstylethree}%

\usepackage[ruled,linesnumbered]{algorithm2e}

\usepackage{algpseudocode}
\algnewcommand\algorithmicforeach{\textbf{for each}}

\begin{document}

\title[BDIS-SLAM]{BDIS-SLAM: A lightweight CPU-based dense stereo SLAM for surgery}


\author*[1,2]{\fnm{Jingwei} \sur{Song}}\email{jingweisong.eng@outlook.com}

\author[2]{\fnm{Ray}~\sur{Zhang}}\email{rzh@umich.edu}

\author[3]{\fnm{Qiuchen} \sur{Zhu}}\email{Qiuchen.Zhu@uts.edu.au}

\author[4]{\fnm{Jianyu} \sur{Lin}}\email{xjtuljy@gmail.com}

\author[2]{\fnm{Maani}~\sur{Ghaffari}}\email{maanigj@umich.edu}

\affil*[1]{\orgdiv{United Imaging Research Institute of Intelligent Imaging}, \orgaddress{\city{Beijing}, \postcode{100144},  \country{China}}}

\affil[2]{\orgname{University of Michigan}, \orgaddress{\city{Ann Arbor}, \postcode{48109}, \state{MI}, \country{USA}}}

\affil[3]{\orgname{University of Technology Sydney}, \orgaddress{\city{Sydney}, \postcode{2007}, \state{NSW}, \country{Australia}}}

\affil[4]{\orgname{Imperial College London}, \orgaddress{\city{London}, \postcode{SW72AZ}, \country{UK}}}




\abstract{\textbf{Purpose:} 
Common dense stereo Simultaneous Localization and Mapping (SLAM) approaches in Minimally Invasive Surgery (MIS) require high-end parallel computational resources for real-time implementation. Yet, it is not always feasible since the computational resources should be allocated to other tasks like segmentation, detection, and tracking. To solve the problem of limited parallel computational power, this research aims at a lightweight dense stereo SLAM system that works on a single-core CPU and achieves real-time performance (more than 30 $Hz$ in typical scenarios). 

\textbf{Methods:} A new dense stereo mapping module is integrated with the ORB-SLAM2 system and named BDIS-SLAM. Our new dense stereo mapping module includes stereo matching and 3D dense depth mosaic methods. Stereo matching is achieved with the recently proposed CPU-level real-time matching algorithm Bayesian Dense Inverse Searching (BDIS). A BDIS-based shape recovery and a depth mosaic strategy are integrated as a new thread and coupled with the backbone ORB-SLAM2 system for real-time stereo shape recovery.

\textbf{Results:} Experiments on in-vivo data sets show that BDIS-SLAM runs at over 30 $Hz$ speed on modern single-core CPU in typical endoscopy/colonoscopy scenarios. BDIS-SLAM only consumes around an additional $12\%$ time compared with the backbone ORB-SLAM2. Although our lightweight BDIS-SLAM simplifies the process by ignoring deformation and fusion procedures, it can provide a usable dense mapping for modern MIS on computationally constrained devices.

\textbf{Conclusion:} The proposed BDIS-SLAM is a lightweight stereo dense SLAM system for MIS. It achieves 30 $Hz$ on a modern single-core CPU in typical endoscopy/colonoscopy scenarios (image size around $640 \times 480$). BDIS-SLAM provides a low-cost solution for dense mapping in MIS and has the potential to be applied in surgical robots and AR systems. Code is available at \url{https://github.com/JingweiSong/BDIS-SLAM}.}




\maketitle

\section{Introduction}

     Simultaneous Localization And Mapping (SLAM) plays a crucial role in modern intra-operative AR \cite{haouchine2013image,widya20193d,portales2022mixed}, Minimally Invasive Surgery (MIS), diseases diagnosis \cite{ratheesh2016advanced, mahmood2019polyp,jia2020automatic} and surgical robotic applications \cite{ma2020visual,ma2022augmented}. SLAM, applied in surgical scenarios, can be classified into two categories based on the mapping manner: sparse and dense. Sparse SLAM focuses on the correct localization of the robot. The two categories of sparse SLAM, Filter-based SLAM~\cite{grasa2011ekf,lin2013simultaneous} and graph-based SLAM~\cite{mahmoud2016orbslam,mahmoud2017slam,turan2017non,oliva2022orb}, all demonstrate satisfying efficiency and robustness in scope localization. Dense SLAM approaches provide additional dense 3D mapping module~\cite{song2017dynamic,song2018mis,zhou2021real,lamarca2020defslam,yang2022endoscope,liu2022sage} and are common in recent applications. Except for DefSLAM \cite{lamarca2020defslam}, which uses monoscope, most works use stereoscope and are deployed on high-end GPUs to ensure real-time colored shape recovery and fusion, providing spatial information among the scope, soft-tissue surface, and surgical instruments. Real-time 3D dense mapping enables surgical robots' 3D perception and ensures dynamic manipulation of surgical instruments, like suturing and cutting. Dense stereo shape recovery algorithms are base modules for high-level mapping or SLAM systems. Among the above-mentioned Stereo SLAM, their mapping module follows the workflow of conducting local stereo shape recovery and then fusing all shapes.\par

    To our knowledge, all dense SLAM in surgical scenarios requires heavy computation on the GPU end, which is not always available. The General-Purpose Graphics Processing Units (GPGPU) mainly handle parallel computations in stereo shape recovery and deformable overlapping shape fusion~\cite{whelan2015elasticfusion}. With the progress of Deep Neural Networks (DNN), GPU is also occupied for other computational-demanding tasks like detection, segmentation, or disease diagnosis~\cite{jia2020automatic}. Integrating more computational devices is difficult due to hardware design constraints in thermal and physical dimensions \cite{song2022bdis}. Therefore, this article proposes a lightweight dense SLAM system that works on the CPU end for the stereoscope.\par

    Specifically, a lightweight dense SLAM in MIS can be achieved by overcoming two challenges: lightweight real-time dense shape recovery from stereo images and non-rigid fusion. Due to the difficulty in fast pixel-wise stereo shape recovery, it is difficult to obtain a real-time CPU-level stereo algorithm that can be used for dense SLAM. Currently, the GPU version of ELAS~\cite{geiger2010efficient} is still widely used in industry~\cite{zampokas2018real,cartucho2020visionblender} and academy~\cite{song2017dynamic,song2018mis,zhang2017autonomous,zhan2020autonomous}. Its CPU version achieves $0.25-1$ second on a single-core modern CPU. A faster 3D shape recovery system is necessary for dense mapping modules at the CPU level. The past few years have witnessed great progress in DNN-based stereo shape recovery algorithms~\cite{chang2018pyramid,yang2019hierarchical,guo2019group,tonioni2019real,xu2020aanet,brandao2020hapnet,long2021dssr}. However, most existing usable DNN-based methods are still supervised and run in near real-time~\cite{allan2021stereo}.\par

    Meanwhile, non-rigid fusion module also presents difficulty in CPU-level dense stereo SLAM. The obtained 3D dense shapes are usually fused in RGB-D style, either in volume-based constrained space \cite{newcombe2011kinectfusion,izadi2011kinectfusion} or surfel-based free space \cite{whelan2015elasticfusion}. Both processes involve massive voxel-wise or surfel-wise fusion calculation. Additionally, non-rigid fusion requires additional time-consuming warping field estimation of the model. The warping field defines the fine-scale model-to-frame deformation field. Obtaining an accurate warping field requires solving a large-scale non-linear objective function~\cite{whelan2015elasticfusion,newcombe2015dynamicfusion,song2018mis}, and it is too expensive on the CPU end.\par

    This research deals with the two challenges, lightweight real-time dense shape recovery, and overlapping shapes fusion, and creates the first lightweight dense stereo SLAM system for the MIS scenario. It leverages our recent Bayesian Dense Inverse Searching (BDIS) algorithm, which is the recent progress in real-time CPU-level stereo dense matching algorithm~\cite{song2022bayesian,song2022bdis}. Researches~\cite{song2022bayesian,song2022bdis} have shown BDIS runs at $14-17$ $Hz$ on $640 \times 480$ stereo images and is slightly more accurate than ELAS. BDIS uses estimated probability to filter unreliable predictions, and its frame rate for stereo shape recovery is sufficient for dense 3D mapping in SLAM systems. BDIS-SLAM simplifies the traditional surfel-based fusion with a rigid mosaic strategy to accelerate the fusion step. The dense stereo shape recovery and 3D mosaic modules have been successfully integrated with the open-sourced ORB-SLAM2 system~\cite{mur2017orb}, which is widely used in the surgical robot community \cite{mahmoud2016orbslam,mahmoud2017slam,mahmoud2018live,song2018mis,oliva2022orb}. It should be pointed out that the mapping accuracy of our lightweight BDIS-SLAM is inferior to other mentioned GPU-based dense SLAM because the warping estimation is omitted in the balance of speed and performance. Overall, this research has the following contributions:\par
    \begin{enumerate}[i.]
		\item To the best of our knowledge, BDIS-SLAM is the first CPU-level dense stereo SLAM for modern MIS.
		\item The recent real-time stereo matching algorithm, BDIS, is coupled with the backbone approach ORB-SLAM2.  
		\item A simplified surfel mapping system is used to be compatible with limited computational resources.
		\item An open-source C++ implementation will be released upon the acceptance.
	\end{enumerate}

	\section{Methodology}
	
	\subsection{Framework overview}
	\label{section_2_1}
	
\begin{figure}[!h]
		\centering
		\subfloat{
			\begin{minipage}[]{1\textwidth}
				\centering
				\includegraphics[width=1\linewidth]{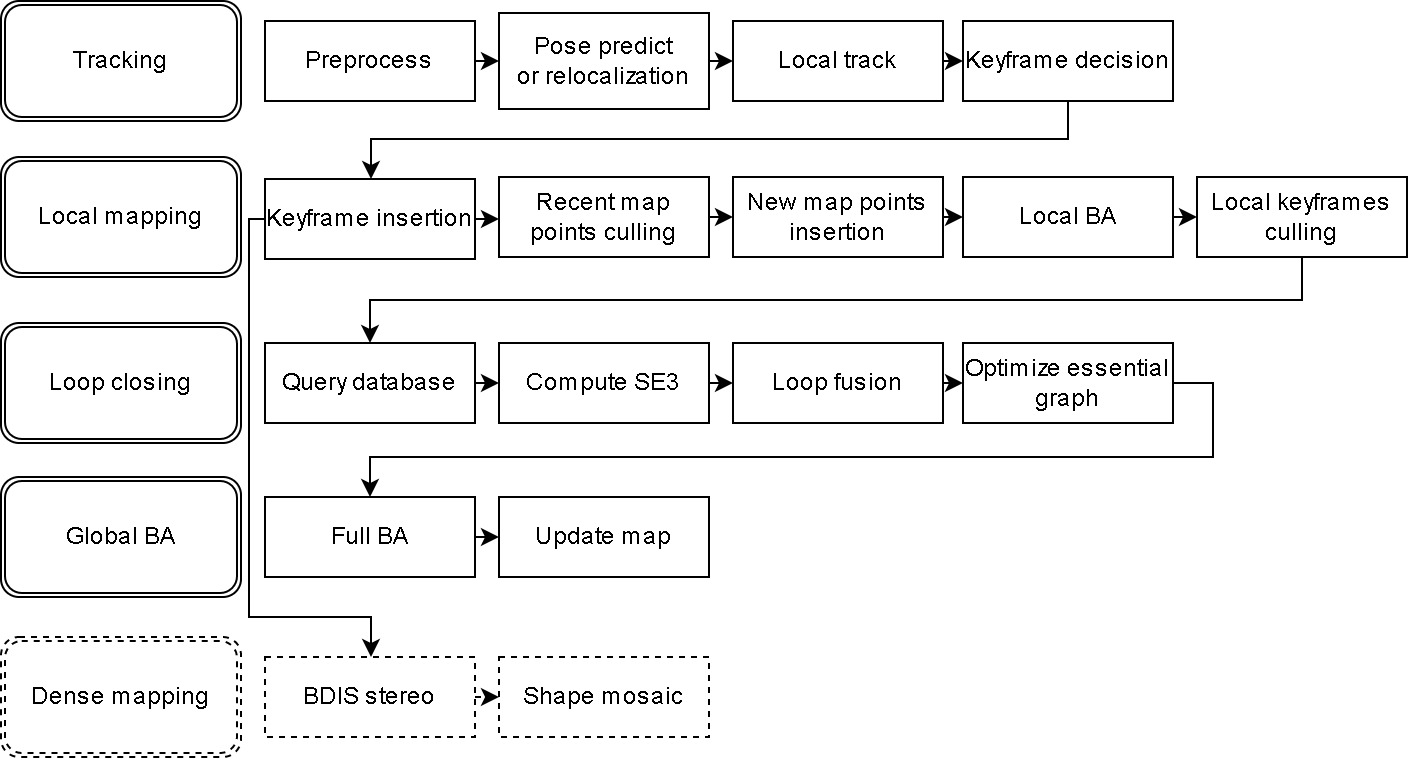}
			\end{minipage}
		}
		\caption{Presented is the framework of BDIS-SLAM. The module in the dashed line is our proposed dense mapping module, while the remaining strictly follows ORB-SLAM2. The arrows show the data flow that begins with stereo data obtaining and preprocessing.}
		\label{fig_framework}
	\end{figure}

	The framework of the proposed BDIS-SLAM is shown in Fig.~\ref{fig_framework}. BDIS-SLAM adopts ORB-SLAM2~\cite{mur2017orb} as the backbone localization method following previous researches~\cite{mahmoud2016orbslam,mahmoud2017slam,mahmoud2018live,song2018mis}. Other stereo sparse SLAM backbones are compatible with our lightweight dense mapping too. In line with~\cite{mahmoud2018live}, the dense mapping module is coupled with ORB-SLAM2 as a new parallel thread. As Fig. \ref{fig_framework} shows, when a new keyframe is inserted into the local mapping system (sparse mapping), our dense mapping module is activated to carry out stereo global shape recovery and model-to-frame global shape mosaic. The shape recovery module implements BDIS to reconstruct a local 3D dense map bounded by the Field of View (FoV) of the captured stereo images. The shape mosaic module fuses the recovered shape and does dense points culling to save memory and computation in the visualization module. DNN-based stereo shape recovery methods cannot be adopted in this study because most of them are still near real-time on GPU and slow on CPU~\cite{song2022bdis}. DNN-based methods still require additional training data set~\cite{allan2021stereo}, which is difficult to obtain. These approaches contradict our requirement for CPU-level lightweight algorithm. Therefore, the applicable stereo-shape recovery methods are prior-free approaches like BDIS.\par 
 
    Although BDIS can process $640\times480$ in $14-17$ $Hz$ on a modern CPU, BDIS-SLAM still follows map-on-keyframe style~\cite{mur2017orb} instead of frame-wise mapping. This is because the model-to-frame fusion (shape mosaic module in this article) requires over $0.5$ second on CPU end after stereo shape recovery and cannot support real-time performance. Map-on-keyframe style~\cite{mur2017orb} saves most computational resources for the backbone SLAM algorithm. Our tests show that ORB-SLAM2 occupies most of the computational resources (our casual estimation indicates around $90\%$), leaving only a limited proportion available for stereo matching. Besides, computational resources must be additionally allocated to multiple shape mosaics and dense mapping visualization in OpenGL, which are also resource-demanding. The map-on-keyframe strategy balances the efficiency and 3D mapping quality.\par

	\subsection{Brief review of BDIS}
	\label{section_2_2}

    The key module in BDIS-SLAM is the fast and accurate BDIS-based stereo shape recovery algorithm~\cite{song2022bayesian,song2022bdis}. Stereo shape recovery is the procedure of triangulating the 3D shape by estimating the parallax between the left and right images. The stereo image pairs are rectified in the camera rectification process, and parallax only exists horizontally~\cite{scharstein2002taxonomy}. BDIS is the current leading robust and fast CPU-level algorithm capable of reconstructing 3D shapes within the FoV. This section provides a brief review of BDIS's concept.\par

    The key idea of BDIS is the fast Lucas-Kanade (LK) algorithm~\cite{baker2004lucas,kroeger2016fast}. Fast LK utilizes a Jacobian trick and is significantly faster than the original LK~\cite{kroeger2016fast}. It searches for the optimal disparity by minimizing

    \begin{equation}
 \label{Eq_fast_inverse_search}
	\Delta \mathbf{u}=\operatorname{argmin}_{\Delta \mathbf{u}^{\prime}} \sum_{x}\left[I_{r}\left(\mathbf{x}+\mathbf{u}\right)-I_{l}(\mathbf{x}+\Delta \mathbf{u}^{\prime})\right]^{2},
	\end{equation}

 \noindent where $\mathbf{x} \in \mathbb{R}^{2\times1}$ is the center of a specific patch, $\mathbf{u}$ denotes the estimated disparity in one loop, $I_l$ and $I_r$ are the left image and right image, $I_r(\cdot)$ and $I_r(\cdot)$ are the left and right patch centering at $\cdot$, and $\Delta \mathbf{u}$ is the optimal update of $\mathbf{u}$ in one loop. Fast LK update $\mathbf{u}$ iteratively with the estimated $\Delta \mathbf{u}^{\prime}$ during the loop. The iteration in \eqref{Eq_fast_inverse_search} is much faster than LK since the Jacobian of the left patch $I_l$ remains fixed in the loop.\par

 BDIS enhances the robustness of fast LK through two strategies~\cite{song2022bayesian}: simplified Conditional Random Fields (sCRF) and Gaussian Mixture Model (GMM). Although~\cite{kroeger2016fast} shows that fast LK achieves real-time processing rate in stereo shape recovery, experiments in~\cite{song2022bayesian} found that it fails to yield satisfactory results. The original fast LK suffers from ambiguities mainly due to three reasons. Firstly, textureless surfaces lead to ambiguous solutions in parallax estimation. This brings about a violation of the photometric consistency, which is the gold standard. Secondly, point-light sources cause severe non-Lambertian reflectance, which is difficult to compensate with models such as the affine formulation used in other fields \cite{engel2017direct}.  This issue severely affects the photometric consistency assumption in cases of small reflection angles. Lastly, dark illuminations, typically caused by large reflection angles, make stereo shape recovery more challenging.\par 
 
 BDIS's pixel-wise probability estimation module enhances the robustness by correctly evaluating overlapping patches' confidences. The patch-wise parallax estimated from \eqref{Eq_fast_inverse_search} is evaluated with an sCRF probability model. The sCRF probability is then distributed pixel-wise based on a GMM model. With the proposed sCRF and GMM formulation, the posterior probability of the patch-wise disparity $\mathbf{u}^{(k)}$ at patch $k$'s center $\mathbf{x}$ is set as\par
 
 \begin{equation}
 \footnotesize 
	\begin{aligned}
	\label{Eq_final_prob}
	p(\mathbf{u}^{(k)}\! \vert \!I_l,I_r,\mathbf{x})\! \propto\! \underbrace{\exp\left(\!-\frac{\!\sum_{\!\mathbf{\xi}^{(k)}(\mathbf{x})}\!\lVert\mathbf{x}-\mathbf{\xi}^{(k)}(\mathbf{x})\lVert^2_2}{2\sigma_s^2}\right)}_{\text{GMM}}
    \underbrace{\frac{\!\exp\left(-\frac{\!\lVert I_l(\mathbf{u}^{(k)})-I_r(\mathbf{u}^{(k)})\!\lVert^2_\mathrm{F}}{2\sigma_r^2\mathrm{s}^2}\!\right)}
	{\!\sum_{\mathbf{u}_i^{(k)}\! \in\! \mathcal{P}}\exp\left(-\frac{ \lVert I_l(\mathbf{u}_i^{(k)})-I_r(\mathbf{u}_i^{(k)})\lVert^2_\mathrm{F}}{2\sigma_r^2\mathrm{s}^2}\!\right)}}_{\text{sCRF}}.
	\end{aligned}
	\end{equation}

    \noindent where $\mathcal{P}$ denotes the domain of all possible choices of $\mathbf{u}_i^{(k)}$, $\lVert\cdot\lVert_\mathrm{F}$ is the Frobenius norm, $\sigma_s$ is the 2D spatial variance of the probability and $\sigma_r$ is the hyperparameter that characterizes the variance of the brightness. $\sigma_r$ is set as the standard deviation of $\lVert I_l(\mathbf{u}^{(k)})-I_r(\mathbf{u}^{(k)})\!\lVert^2_\mathrm{F}$. $\mathbf{\xi}^{(k)}(\mathbf{x})$ refers to the set of all pixel positions within the patch $k$ in the image coordinate. In the sCRF term, sampling is performed within a window. In practice, the sampling step in sCRF's term is simplified to a small, fixed small window, and the impact of the remaining candidates is assumed to be negligible. In the GMM term, the patch probability is applied to each pixel proportionally. Since the GMM term is independent of the pixel position in image $\mathbf{u}^{(k)}$, it can be precomputed. The probability is used as a weight for overlapping patches and also as a threshold for filtering unreliable predictions. Qualitative and quantitative comparisons with ELAS and SGBM in~\cite{song2022bayesian,song2022bdis} show that local minima and saddle point solutions have been reduced significantly. With the Bayesian-based probability estimation algorithm, BDIS achieves slightly better accuracy than near real-time ELAS~\cite{song2022bayesian}. The pixel-wise optimal parallax $\hat{\mathbf{u}}_\mathbf{x}$ is fused as\par 

    \begin{equation}
	\label{Eq_residual_fusion}
	\hat{\mathbf{u}}_\mathbf{x} = \sum_{k \in \Omega} \frac{ p(\mathbf{u}^{(k)} \vert I_l,I_r,\mathbf{x})}{\sum_{k \in \Omega} p(\mathbf{u}^{(k)} \vert \!I_l,I_r,\mathbf{x})}  \mathbf{u}^{(k)},
	\end{equation}
	\noindent where $\Omega$ is the set of patches covering the 2D position $\mathbf{x}$. The pixel-wise disparity $\hat{\mathbf{u}}_\mathbf{x}$ is the weighted average of the estimated disparities from all patches, wherein the weight is the inverse residual of brightness.\par

	\subsection{Mapping and 3D mosaic procedure}
	\label{section_2_3}
	
	This section elucidates the dense mapping thread's structure depicted in Fig.~\ref{fig_framework}. Whenever a keyframe is inserted, its stereo image pair is used to retrieve the disparity in \eqref{Eq_residual_fusion} in the finest scale denoted as $\hat{\mathbf{u}}_\mathbf{x}^{(f)}$($(f)$ is the finest scale and $\mathbf{x}$ is thus the pixel location). Subsequently, the depth is computed following the routine stereo shape recovery process, which is\par  

    \begin{equation}
	\label{Eq_depth_fusion}
	\mathrm{d}_\mathbf{x} = \frac{\mathrm{f}\mathrm{b}}{\hat{\mathbf{u}}_\mathbf{x}^{(f)}},
	\end{equation}
	\noindent where $\mathrm{f}$ and $\mathrm{b}$ represent the focal length and baseline provided by stereo camera calibration. The 3D point in world coordinate $\mathbf{p}^{(i)}_x \in \mathbb{R}^{4\times1}$ (in homogeneous coordinate) in keyframe $i$ is\par

    \begin{equation}
    \label{eq_3d_pts_lift}	\mathbf{p}^{(i)}_x=\mathbf{T}^{(i)}\mathbf{K}^{-1}[\mathbf{x}\mid \mathrm{d}_x \mid 1],
	\end{equation}

    \noindent where $\mathbf{T}^{(i)} \in \mathrm{SE}(3)$ is the pose of the camera defined as camera-to-world transformation, $\mathbf{K} \in \mathbb{R}^{3\times4}$ is the camera intrinsic matrix and $\mathbf{K}^{-1}$ denotes the pseudo inverse of $\mathbf{K}$. By deploying \eqref{eq_3d_pts_lift}, the point cloud set $\mathbf{P}^{(i)}=\left\{\mathbf{p}^{(i)}_x, \mathbf{x} \in \operatorname{R}(I_l) \right\}$ is retrieved within the FOV, where $\operatorname{R}(\cdot)$ defines the range of the depth's 2D coordinate. The dense mapping thread's details are presented in Algorithm \ref{Algorithm_bdis_thread}.\par

    In contrast to prior works that utilize volume and surfel-based fusion, the proposed approach in this study simplifies the process by mosaicking all $\mathbf{P}^{(i)}$. There are three reasons. First, the restricted hardware does not support expensive warping field estimation. Estimating an optimal warping field requires solving a non-linear objective function in large size and large state space, which is not possible with current hardware. Secondly, BDIS-SLAM fuses local shapes in a map-on-keyframe manner rather than frame-wise fusion, as seen in earlier studies (e.g., \cite{whelan2015elasticfusion,newcombe2015dynamicfusion,song2018mis}). The keyframe-wise fusion is restricted by the limited number of observations, leading to unreliable correction of outliers and errors based on the few overlapping patches. Lastly, without accurate deformation field and outlier rejection, volume or surfel base methods yield blurry shapes by blending the texture of multiple overlapping shapes. In contrast, the dense map is presented in the form of raw colored points, which consist of positions $\mathbf{p}^{(i)}_x$ and color. BDIS-SLAM scarifies the accuracy of fusion and just mosaic all recovered shapes. The previously recovered shape is substituted in regions that overlap with the newly retrieved shape. In summary, BDIS-SLAM sacrifice dense mapping accuracy in exchange for real-time performance on CPU end.\par 
	
	\begin{algorithm}[t]
	\caption{keyframe-wise shapes mosaic.}
	\label{Algorithm_bdis_thread}
	\KwIn{Stereo image at frame $i$, $\mathbf{K}$, camera pose at $i$th frame $\mathbf{T}^{(i)} \in \mathrm{SE}(3)$, recovered depth map at $i$th frame $\mathbf{D}^{(i)}$, key frame index start from $1$}
	\KwOut{Fused dense map $\mathbf{P}$}
	$r=1$\;
	\While{$r=i$}{
        Obtain $\mathbf{P}^{(i)}$ with BDIS, \eqref{Eq_depth_fusion} and \eqref{eq_3d_pts_lift}\;
		\eIf{$r = 1$}{
            $\mathbf{P}={\{\mathbf{P}^{(i)}\}}$\;
        }{
            \For{$\mathbf{p}_x \in \mathbf{P}$}{
        		/* Check if 2D projection within current depth's range */  \;
                \If{$\mathbf{K}{\mathbf{T}^{(i)}}^{-1}\mathbf{p}_x \in \operatorname{R}(\mathbf{D}^{(i)})$}
                {
                    Delete $\mathbf{p}_x$ in $\mathbf{P}$\;
                }
            }
            $\mathbf{P} = \mathbf{P} \cup \mathbf{P}^{(i)}$\;
            Send $\mathbf{P}$ for visualization \;
    		$r=r+1$\;
        }
	}
	Notation: $\operatorname{R}(\cdot)$ defines the range of the depth's 2D coordinate.\\
\end{algorithm}

\section{Results and discussion}

    We evaluate the proposed SLAM from three aspects: time consumption, visual appearance, and the accuracy of virtual data sets. To our knowledge, there is no similar CPU-level lightweight dense stereo SLAM for stereoscopes similar to BDIS-SLAM. Thus, the comparison with other researches on in-vivo data sets is unavailable. We emphasize that we DO NOT claim BDIS-SLAM to be more accurate than other GPU-based baselines. The localization accuracy of similar stereo SLAM \cite{mahmoud2016orbslam,song2018mis} depends on the backbone (ORB-SLAM2 used in most approaches); the mapping accuracy depends on the stereo matching algorithm. Our preliminary stereo matching algorithms \cite{song2022bayesian,song2022bdis} provide comprehensive quantitative and qualitative comparisons on stereo shape recovery accuracy, speed, and ablation studies.\par 

    Consequently, our experiments focus on validating our core proposition of BDIS-SLAM as a practical and time-efficient lightweight dense SLAM system. We first show that the dense mapping module adds a limited amount of computational burden to its backbone ORB-SLAM2. Then, the 3D dense mapping results of BDIS-SLAM are visually demonstrated. Lastly, a quantitative test on a synthetic data set is provided.\par

    The proposed BDIS-SLAM was developed based on the open-sourced code ORB-SLAM2 \cite{mur2017orb}, and BDIS \cite{song2022bdis}. The computations were implemented on a commercial laptop ALIENWARE M17 R4 (i7-10870H and 32G RAM) in C++. Our tests indicate that less than 5GB of memory is used on all data sets. All parameter settings of BDIS follow the original work \cite{song2022bdis}. Bayesian window was 5; the disturbance from the convergence was $0.5$ and $1$ pixel; minimal ratio of the valid patch was set to $0.75$; $\sigma_s$ was set to $4$. Pixel-wise threshold $0.15$ is enforced to filter unreliable predictions. The parameter setting of the backbone ORB-SLAM2 follows \cite{mahmoud2016orbslam,song2018mis}.\par
    
    The in-vivo data set with accurate calibration was collected by Hamlyn Centre for Robotic Surgery \cite{giannarou2013probabilistic}. The stereo videos were collected when the endoscopy/laparoscopy procedure was manipulated by the surgeon on the live porcine. The frame rate and image sizes are ($30 Hz$, $[640 \times 480]$ for data set 6) and ($25 Hz$, $[ 720 \times 288]$ for data set 15, 16, 20, 21, 25 and 26). The rest frame cannot be tested due to fast motion, static position, no planar motion for SLAM initialization, lack of stereo images and very short sequence. Additionally, a synthetic colon's stereo video is utilized for quantitative comparison. The virtual colon was used in BDIS research \cite{song2022bayesian,song2022bdis} (Unity3D\footnote{\url{https://unity.com/}}). In this research, the textureless surface has to be substituted with a more ``textured'' surface to ensure the backbone SLAM can locate the virtual scope correctly. The point-to-plane accuracy serves as a performance metric for evaluating the BDIS-SLAM's performance. Overall, there are 12000 consecutive stereo images with accurate camera parameters, reference shapes, and trajectories.\par

	\begin{figure}[htpb]    
	\centering
	\subfloat{    
		\begin{minipage}[htpb]{0.95\textwidth}    
			\centering
			\includegraphics[width=1\linewidth]{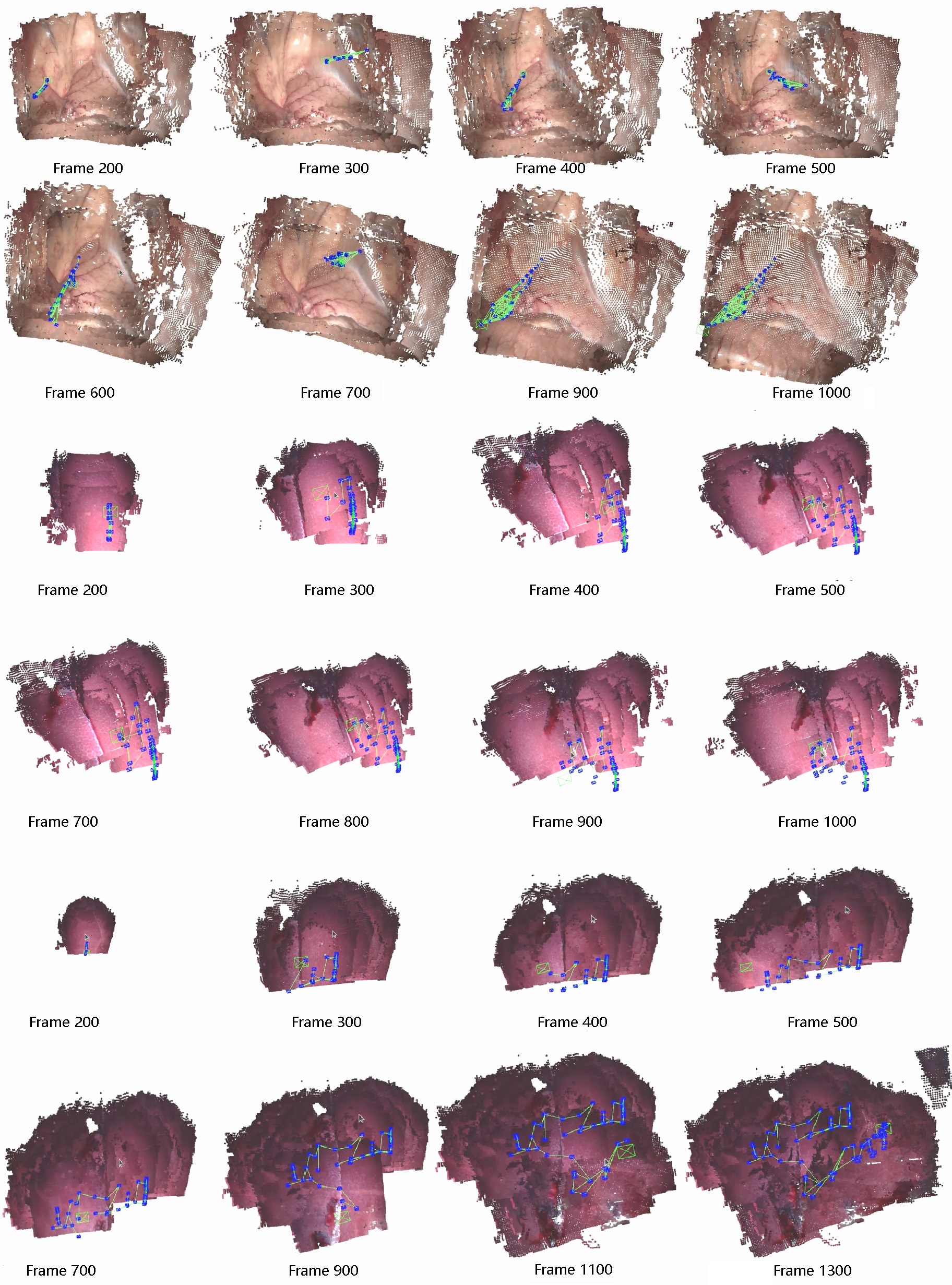}            
		\end{minipage}                
	}\\
	\caption{Sample 3D reconstructions on the three in-vivo stereo endoscopy/laparoscopy data sets by implementing BDIS-SLAM. Rows 1-2, 3-4, and 5-6 show the results on Hamlyn data sets 6, 15, and 16 consecutively.}
	\label{fig_invivo_results}
\end{figure}

\begin{figure}[htpb]    
	\centering
	\subfloat{    
		\begin{minipage}[htpb]{1\textwidth}    
			\centering
			\includegraphics[width=1\linewidth]{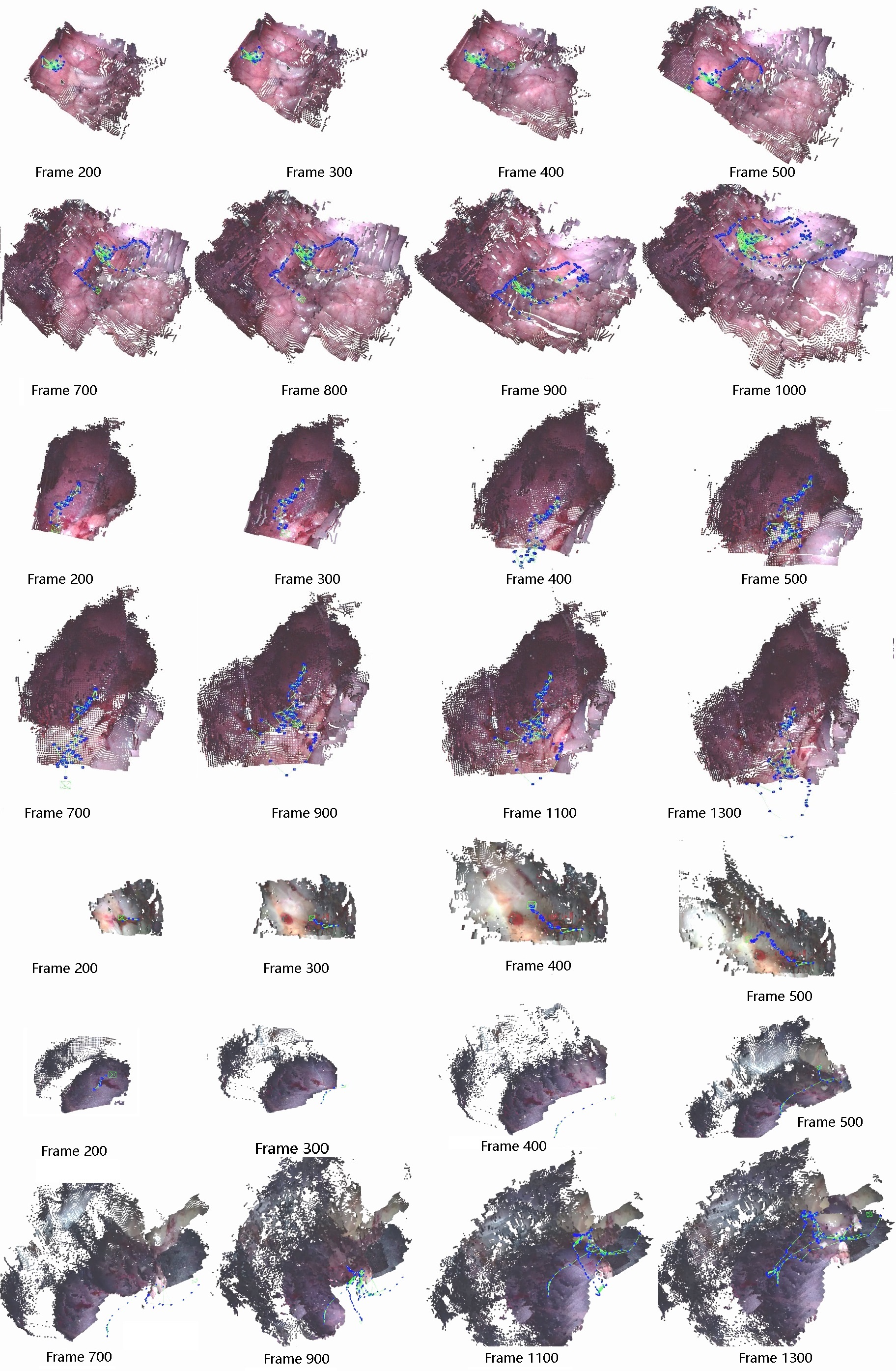}            
		\end{minipage}                
	}\\
	\caption{Sample 3D reconstructions on the three in-vivo stereo endoscopy/laparoscopy data sets by implementing BDIS-SLAM. Rows 1-2, 3-4, 5, and 6-7 show the results on Hamlyn data sets 20, 21, 25 and 26 consecutively.}
	\label{fig_invivo_results_1}
\end{figure}

\begin{figure}[htpb]    
	\centering
	\subfloat{    
		\begin{minipage}[htpb]{1\textwidth}    
			\centering
			\includegraphics[width=1\linewidth]{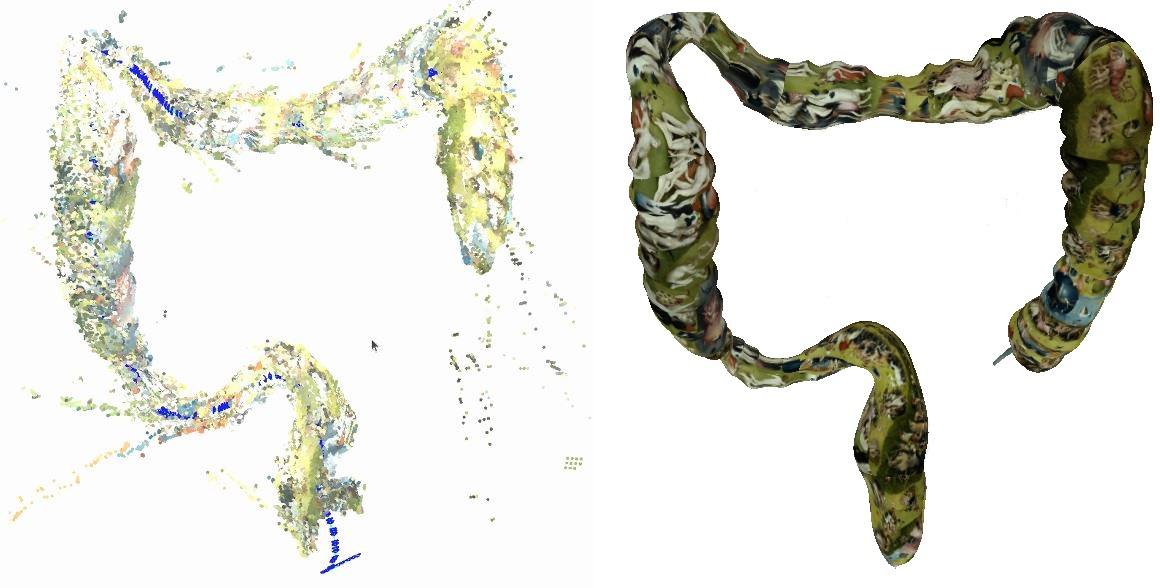}            
		\end{minipage}                
	}\\
	\caption{3D reconstruction on the synthetic data set. Left is the full 3D recovered map of BDIS-SLAM. Right is the original 3D mesh for synthetic endoscopy data simulation.}
	\label{fig_synth_results}
\end{figure}

 \subsection{Reconstruction results}

The qualitative mapping results are presented with sequential 3D models. Fig. \ref{fig_invivo_results} and Fig. \ref{fig_invivo_results_1} shows the sequential qualitative 3D mapping results of the proposed BDIS-SLAM on three Hamlyn in-vivo data set sequences. Fig. \ref{fig_synth_results} shows the final reconstruction on the synthetic colon data set. The attached video also provides a visual illustration. The above results demonstrate the qualitative results of all sequential 3D shapes on a large scale. Fig. \ref{fig_invivo_results}, Fig. \ref{fig_invivo_results_1} and Fig. \ref{fig_synth_results} confirm that stereo shape recovery method BDIS~\cite{song2022bayesian,song2022bdis} is robust in recovering most shape correctly on the tested data sets. Unlike SLAM systems~\cite{song2017dynamic,song2018mis,zhou2021real,yang2022endoscope} which rectify depth error by fusing overlapping depth sequentially, mapping module in BDIS-SLAM requires more robust stereo shape recovery method since only shape mosaic strategy is used in Algorithm \ref{Algorithm_bdis_thread}. Results show that BDIS achieves high quality for shape mosaic.\par

Fig. \ref{fig_invivo_results}, Fig. \ref{fig_invivo_results_1} and Fig. \ref{fig_synth_results} also reveal some inherent limitations in the performance of BDIS-SLAM. First, there are noticeable gaps in the mosaic. Since no overlapping shape fusion module is provided in Algorithm \ref{Algorithm_bdis_thread}, the gap is inevitable. Second, texture illuminations are different. In the in-vivo data collection process, the activated illumination brings serious non-Lambertian reflectance problems, which can hardly be solved in a computation-constrained scenario. Lastly, BDIS-SLAM suffers from the influence of deformation. Previous work modeling deformation by solving a high dimensional deformation field on GPU-end while BDIS-SLAM cannot afford such computation.\par

We also quantitatively tested the reconstruction results in Fig. \ref{fig_synth_results}. Average mean and median point-to-point distances between the reconstructed map and the original mesh model were calculated as

\begin{align}
	\label{Eq_abs_error}
	\mathrm{e}_x=\lVert \mathbf{p}_x - \overline{\mathbf{p}_x} \lVert_2,
\end{align}

\noindent where $\overline{\mathbf{p}_x}$ is the closest point on original mesh model. Outliers (error distances larger than $5 mm$) were excluded from this test. If $\mathrm{e}_x$ is less than $5 mm$, $\mathrm{e}_x$ is added as a new element to error set as

\begin{align}
	\label{Eq_abs_error_1}
    E = E \cup \mathrm{e}_x,
\end{align} 

\noindent where $E$ is the set of errors and initialized as an empty set. Median and mean errors are calculated based on $E$. Results show that BDIS-SLAM achieves average mean and median errors $0.296 mm$ and $0.215 mm$, which is similar to the stereo shape recovery accuracy (mean and median $0.202 mm$ and $0.161 mm$) reported in BDIS~\cite{song2022bdis}. It should be emphasized that quantitative comparison with other approaches is not achievable and necessary. No similar CPU-based stereo SLAM algorithm can be found. The closest method~\cite{mahmoud2018live} does not reveal the code and their Zero Mean Normalized Cross Correlation based stereo matching algorithm, similar to SGBM, has been demonstrated to be inferior to BDIS~\cite{song2022bayesian,song2022bdis}. \textbf{The accuracy of the dense mapping module is predominantly dependent on stereo matching algorithm which has been extensively validated in BDIS~\cite{song2022bayesian,song2022bdis}. }\par

\subsection{Time consumption}

The major benefit of BDIS-SLAM lies in the balance of time consumption and performance. We tested BDIS-SLAM on three in-vivo data sets and recorded the frame rates. The images of all sequences are fed into the system on demand and their actual frame rates are ignored, thus revealing the maximum processing capability of BDIS-SLAM. Table \ref{Table_exvivo_dataset} presents the processing rate of BDIS-SLAM and its backbone ORB-SLAM2. As BDIS-SLAM is a combination of ORB-SLAM2 and a dense mapping thread, table \ref{Table_exvivo_dataset} shows the additional computation burden of this thread. It reveals that the dense mapping thread gains around $12\%$ additional burden for the three in-vivo data sets, which remains within acceptable limits as they are still over $30 Hz$. Overall, BDIS-SLAM consumes double the computation than its backbone ORB-SLAM2~\cite{mur2017orb}. There are two major reasons: First, the reprojection iterative process ($\mathbf{P}$ size in Algorithm \ref{Algorithm_bdis_thread}) is heavy when the scenario is large. Second, visualization on OpenGL also requires traversing $\mathbf{P}$. Both procedures are linearly correlated to size $\mathbf{P}$. Fortunately, most MIS applications only work on small space instead of the entire colon shown in Fig. \ref{fig_synth_results}.\par

Similar CPU-level reconstruction work can be found in~\cite{mahmoud2018live}, which also performs stereo shape recovery. While the frame rate is not reported, \cite{mahmoud2018live} provides a detailed breakdown of time consumption for each stage of the process. Specifically, the authors reported that 3D reconstruction took more than 10 seconds on the CPU end, with the majority of time spent on stereo shape recovery. Take a $720 \times 288$ stereo image (same data collector as data sets 20 and 21 used in our experiments) as an example, the cost volume calculation and variational minimization are $3.4s$ and $6.2s$, respectively. This volume-based aggregation approach calculates the entire cost volume and searches for the best disparity. 

\begin{table}[]
\centering
\caption{The frame rates tested on the three in-vivo data sets and the synthetic data set. All values are in $Hz$. }
\begin{tabular}{lrr}
\hline
            & BDIS-SLAM & ORB-SLAM2 \\ \hline
Data set 6  & 34.42     & 38.53    \\
Data set 15 & 33.09     & 36.45    \\
Data set 16 & 33.15     & 37.00    \\
Data set 20 & 32.65     & 36.04    \\
Data set 21 & 26.94     & 30.46    \\
Data set 25 & 36.59     & 40.65    \\
Data set 26 & 27.93     & 30.96    \\
Synthetic   & 14.41     & 30.80    \\ \hline
\end{tabular}
\label{Table_exvivo_dataset}
\end{table}

	\section{Conclusion}
	
	In this study, we extended the recently proposed CPU-level real-time stereo shape recovery method BDIS and proposed the first CPU-level dense stereo SLAM for modern surgery. A new dense mapping thread is integrated with the backbone ORB-SLAM2 and performs stereo shape recovery and shapes mosaic. Experimental results show that BDIS-SLAM successfully balances time consumption and performance on a single CPU. BDIS-SLAM achieves 30 $Hz$ on a modern single-core CPU in typical endoscopy/colonoscopy scenarios (image size around $640 \times 480$). The lightweight BDIS-SLAM provides a low-cost solution to MIS applications, especially when GPU resources are occupied for tasks like detection, segmentation, or disease diagnosis. Future works may focus on addressing some of the limitations of BDIS-SLAM, like reducing shape gaps, texture illumination differences, and outliers. Moreover, a lightweight fusion process using the predicted depth variance from BDIS can greatly benefit the 3D mapping module.

\section{Declarations}
\subsection{Conflict of interest}
The authors declare no conflict of interest.
\subsection{Ethical approval}
This article does not contain any studies with human participants performed by any of the authors.
\subsection{Informed consent}
This article does not contain patient data.

\bibliography{bib/strings-abrv,bib/ieee-abrv,endoscope}


\end{document}